\documentclass[10pt,twocolumn,letterpaper]{article}

\usepackage[pagenumbers]{cvpr} % To force page numbers, e.g. for an arXiv version

\usepackage[accsupp]{axessibility}
\usepackage{bbm}
\usepackage{amssymb,mathtools}
\usepackage{multirow}
\usepackage{colortbl}
\usepackage{soul}
\usepackage{comment}
\usepackage{makecell}
\usepackage{pifont}
\usepackage{wrapfig}
\usepackage{subcaption}
\usepackage{arydshln}

\newcommand{\cmark}{\ding{51}}%

\definecolor{baselinecolor}{gray}{.93}
\newcommand{\baseline}[1]{\cellcolor{baselinecolor}{#1}}

\definecolor{cvprblue}{rgb}{0.21,0.49,0.74}
\usepackage[pagebackref,breaklinks,colorlinks,allcolors=cvprblue]{hyperref}

\newcommand{\modelname}[0]{DiGIT}
\newcommand{\encodername}[0]{MDGE}
\newcommand{\decodername}[0]{CAID}

\title{\vspace{-2pt}\modelname{}: Multi-Dilated Gated Encoder and Central-Adjacent Region Integrated Decoder for Temporal Action Detection Transformer\vspace{-5pt}}

\newcommand{\authorskip}{\hspace{7.5mm}}
\newcommand{\customfootnote}[1]{%
  \begingroup
  \renewcommand{\thefootnote}{}
  \footnote{#1}%
  \endgroup
}

\author{
Ho-Joong Kim \authorskip Yearang Lee \authorskip Jung-Ho Hong \authorskip
 Seong-Whan Lee$^*$ \\[2.5pt]
 \normalsize Dept. of Artificial Intelligence, Korea University, Seoul, Korea\\[-1pt]
 {\tt\small \{hojoong\_kim, yr\_lee, jungho-hong, sw.lee\}@korea.ac.kr} \vspace{-6pt}\\
}

\begin{document}
\maketitle
\begin{abstract}
In this paper, we examine a key limitation in query-based detectors for temporal action detection (TAD), which arises from their direct adaptation of originally designed architectures for object detection. Despite the effectiveness of the existing models, they struggle to fully address the unique challenges of TAD, such as the redundancy in multi-scale features and the limited ability to capture sufficient temporal context. To address these issues, we propose a multi-dilated gated encoder and central-adjacent region integrated decoder for temporal action detection transformer (DiGIT). Our approach replaces the existing encoder that consists of multi-scale deformable attention and feedforward network with our multi-dilated gated encoder. Our proposed encoder reduces the redundant information caused by multi-level features while maintaining the ability to capture fine-grained and long-range temporal information. Furthermore, we introduce a central-adjacent region integrated decoder that leverages a more comprehensive sampling strategy for deformable cross-attention to capture the essential information. Extensive experiments demonstrate that DiGIT achieves state-of-the-art performance on THUMOS14, ActivityNet v1.3, and HACS-Segment. Code is available at: \href{https://github.com/Dotori-HJ/DiGIT}{https://github.com/Dotori-HJ/DiGIT}
\end{abstract}
\vspace{-22pt}
\customfootnote{*Corresponding author}
\section{Introduction}
\label{sec:intro}
Temporal action detection (TAD) is crucial for video understanding and supports a wide range of real-world applications, including video surveillance, summarization, and retrieval.
TAD aims to detect action instances within untrimmed videos by identifying action classes along with their start and end times.
The most existing TAD methods~\cite{TadTR,TE-TAD,ActionFormer,TriDet} are a snippet-based approach, which utilizes pre-extracted features to address long durations.
This approach does not require the computational cost for the backbone network at the detection stage, enabling the detector to address a comprehensive length of features at once.
The existing methods can be divided into three approaches: anchor-based~\cite{BSN,BMN,BC-GNN,GTAN,G-TAD,VSGN}, anchor-free~\cite{AFSD,TALLFormer,ActionFormer,TriDet,VideoMambaSuite,ASL,DyFADet}, and query-based~\cite{RTD-Net,ReAct,TadTR,TE-TAD,DualDETR}.

\begin{figure}[!t]
    \centering
    \includegraphics[width=0.86\linewidth]{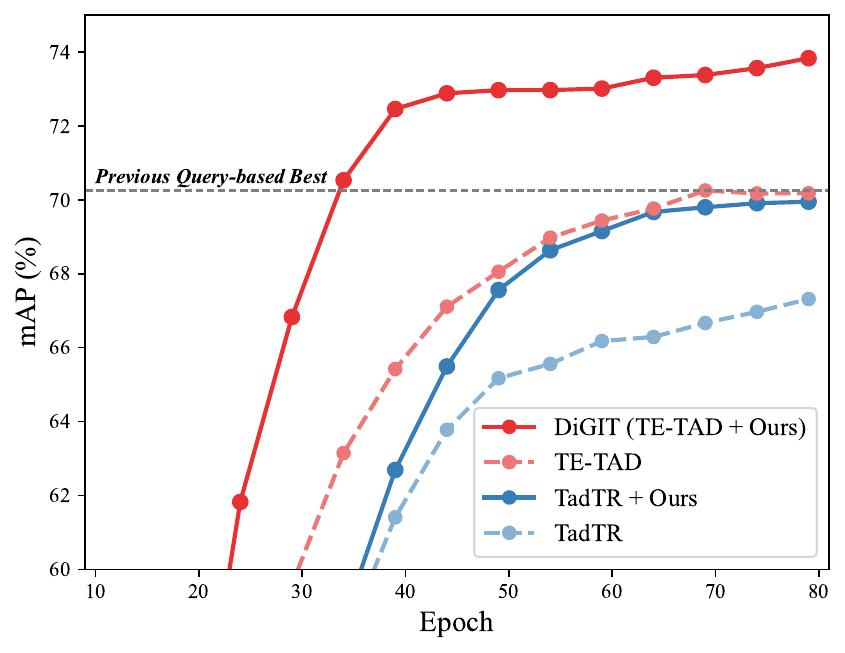}\\ 
    \vspace{-10pt}
    \caption{\textbf{Convergence curves with InternVideo2~\cite{InternVideo2} features on THUMOS14~\cite{THUMOS14}.} Our method boosts the previous query-based detectors like TE-TAD~\cite{TE-TAD} and TadTR~\cite{TadTR}.}
    \vspace{-17pt}
    \label{fig:convergence}
\end{figure}
Query-based detectors, inspired by DETR~\cite{DETR}, have attracted interest because of their potential to eliminate reliance on hand-crafted components, such as the sliding window and non-maximum suppression (NMS).
This capability derives from their adoption of a set-prediction mechanism, which enables an end-to-end detection process through a one-to-one matching paradigm.
Among them, TE-TAD~\cite{TE-TAD} enables a full end-to-end detection process by reformulating the coordinate representation based on recent DETR-based network architectures, such as multi-scale deformable attention~\cite{Deformable-DETR}.
However, despite these advancements, query-based detectors still rely on originally designed architectures for object detection, which is not fully suited to addressing the unique challenges of TAD.
We identify two limitations within the encoder and decoder structures of existing query-based TAD models.
(1) In the encoder, simply utilizing single-scale~\cite{TadTR,DualDETR} feature or multi-scale~\cite{TE-TAD} features fails to extract the meaningful features needed to capture the diverse temporal scale information and distinct feature representations.
(2) In the decoder, the deformable cross-attention mechanism focuses on the central regions of reference points, overlooking surrounding areas essential for accurately detecting the action instances.

\begin{figure}[!t]
    \centering
    \begin{subfigure}[t]{0.9\linewidth}
        \centering
        \hfill
        \includegraphics[width=0.4\linewidth]{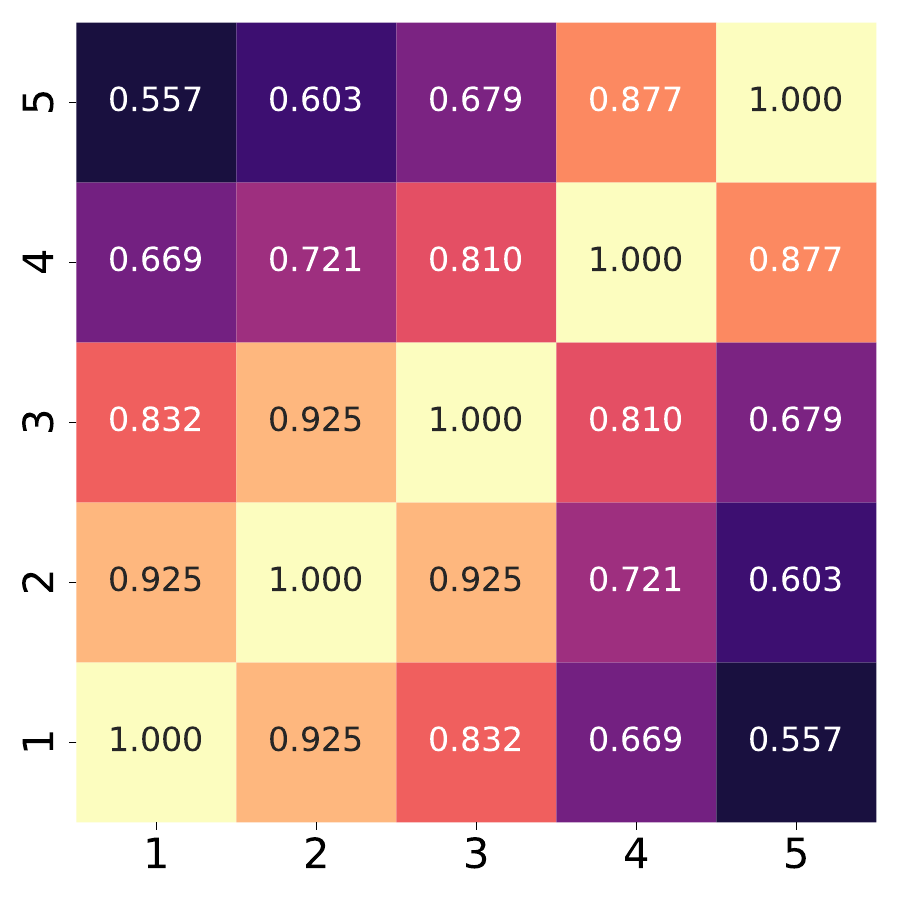}
        \hfill
        \hfill
        \includegraphics[width=0.4\linewidth]{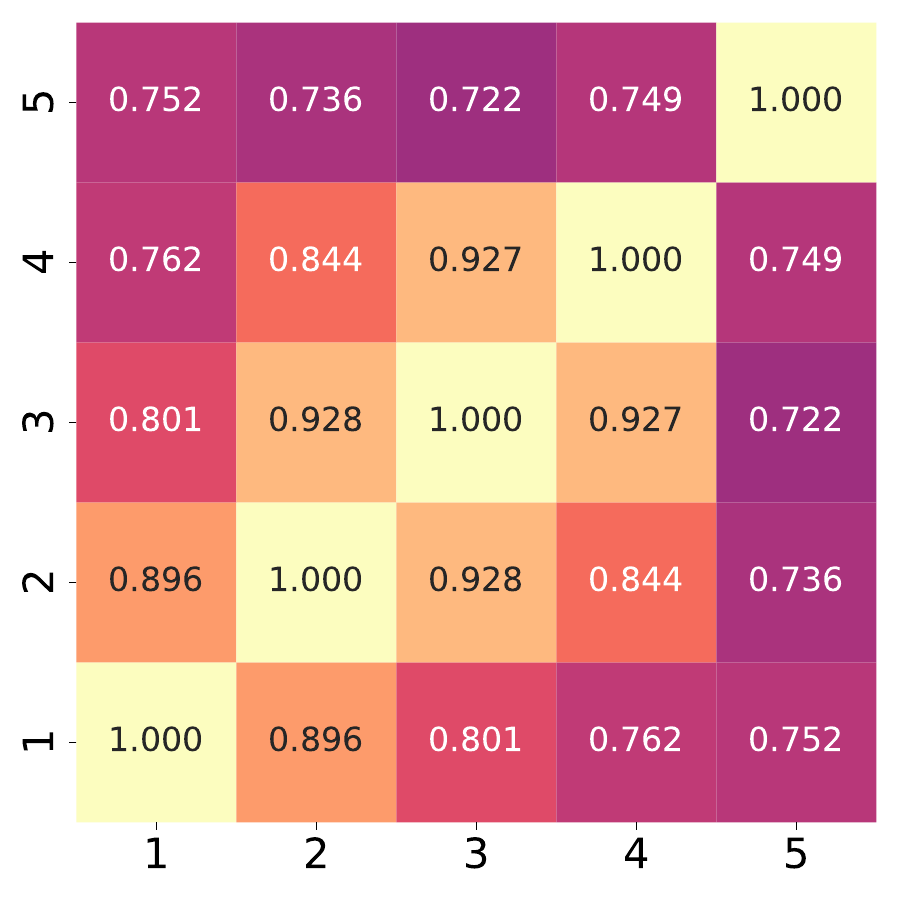}
        \hfill
        \vspace{-3pt}
        \caption{DINO-5scale~\cite{DINO} with ResNet-50~\cite{ResNet} on COCO~\cite{COCO}}
        \label{fig:cka_dino}
    \end{subfigure}%
    \hfill
    \begin{subfigure}[t]{0.9\linewidth}
        \centering
        \hfill
        \includegraphics[width=0.4\linewidth]{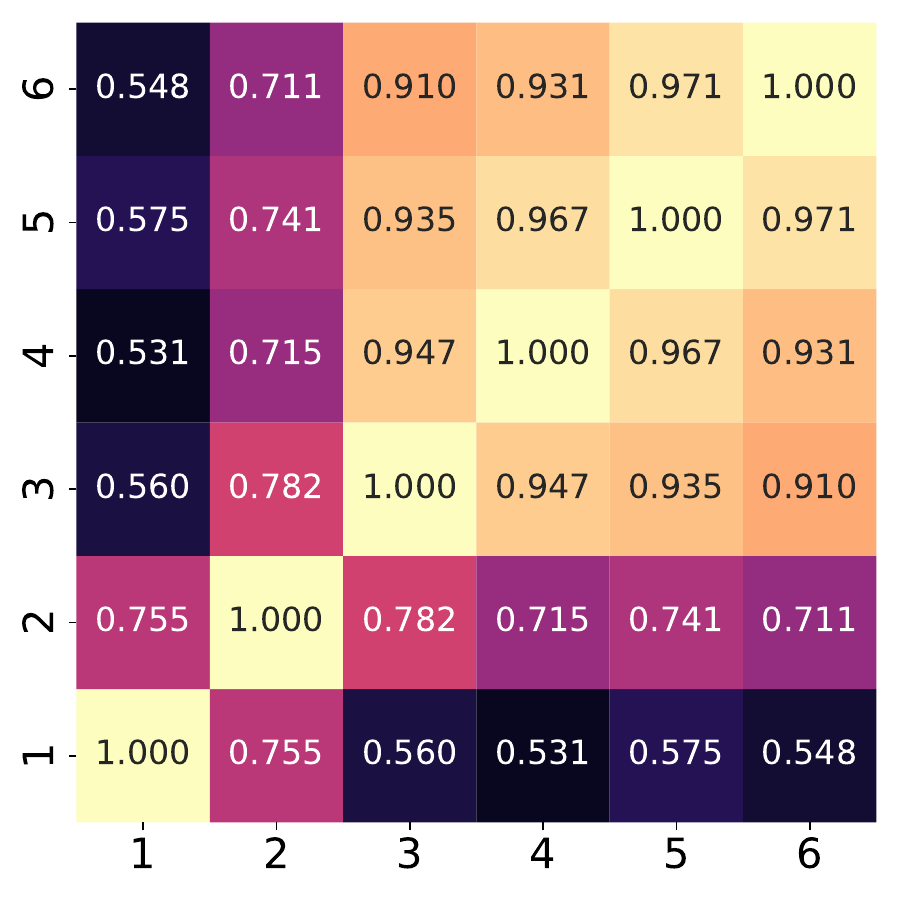}
        \hfill
        \hfill
        \includegraphics[width=0.4\linewidth]{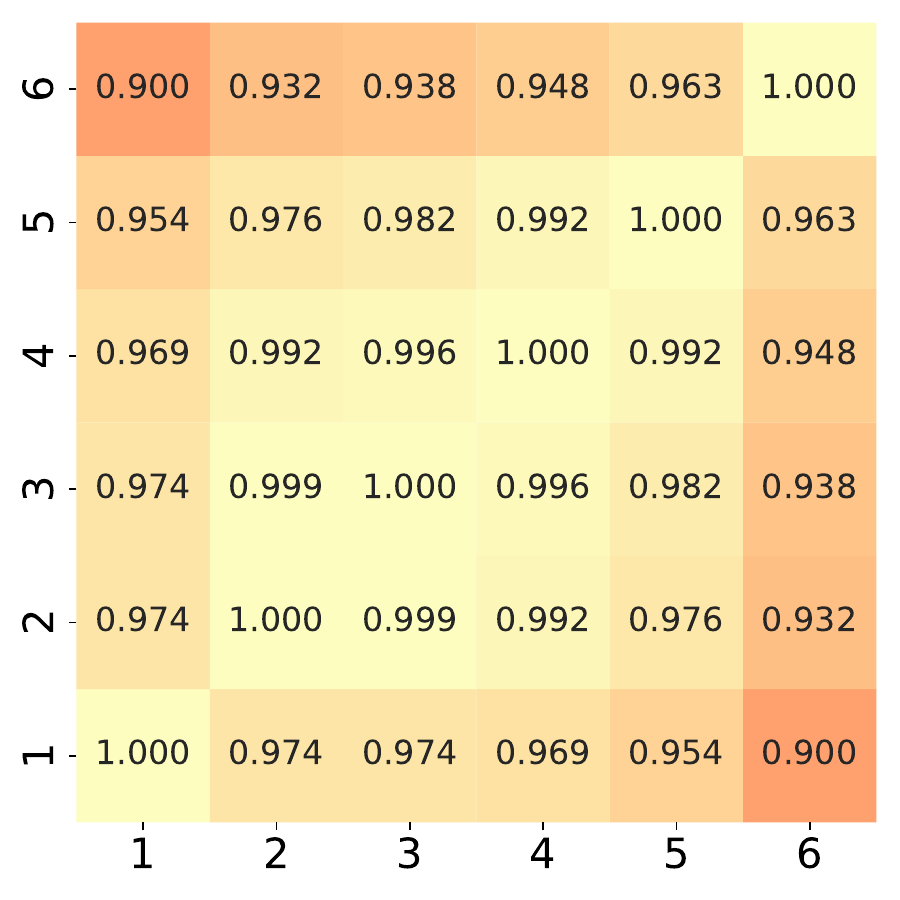}
        \hfill
        \vspace{-3pt}
        \caption{TE-TAD~\cite{TE-TAD} with InternVideo2~\cite{InternVideo2} on THUMOS14~\cite{THUMOS14}}
        \label{fig:cka_te_tad}
    \end{subfigure}

    \vspace{-8pt}
    \caption{\textbf{Layer-wise CKA similarity comparison between object detection and TAD.} The left and right sides are extracted from pre-encoder and post-encoder features, respectively. The 1–5 or 1–6 labels on each axis correspond to the number of multi-scale feature levels used in the respective models.}
    \vspace{-9pt}
    \label{fig:cka}
\end{figure}

In the encoder, existing query-based detectors employ either single-scale~\cite{TadTR,DualDETR} or multi-scale~\cite{TE-TAD} features, but both approaches have inherent limitations.
The single-scale approach~\cite{TadTR,DualDETR} processes feature at a single resolution throughout the encoder and decoder, intuitively restricting the model to capture varying durations.
The multi-scale approach~\cite{TE-TAD} combines features across multiple resolutions, enhancing the ability of the model to detect different lengths of actions by aggregating broader contextual information.
However, despite the advantage of the multi-scale approach, it struggles to capture distinct feature representations at each level, resulting in highly correlated features across scales.
Fig.~\ref{fig:cka} illustrates this issue by comparing layer-wise CKA~\cite{CKA} similarities on the pre-encoder and post-encoder features between an object detection model (DINO~\cite{DINO}) and a TAD model (TE-TAD~\cite{TE-TAD}).
As shown in Fig. \ref{fig:cka} on the left, the pre-encoder features of TE-TAD show high similarity among high-level features (3-6) compared to DINO.
This is due to the repeated use of single convolutional projections, where the final-layer feature is downsampled to produce multi-scale features.
These highly similar pre-encoder features of TE-TAD cause the multi-scale deformable attention to propagate redundant information across levels during encoding.
Consequently, as shown in Fig.~\ref{fig:cka}\subref{fig:cka_te_tad} on the right, the post-encoder features show high similarity across levels compared to DINO.
This result suggests that utilizing multi-scale features from the initial stage leads to excessive redundancy.

\begin{figure}[!t]
    \centering
    \begin{subfigure}[t]{\linewidth}
        \centering
        \includegraphics[width=\linewidth]{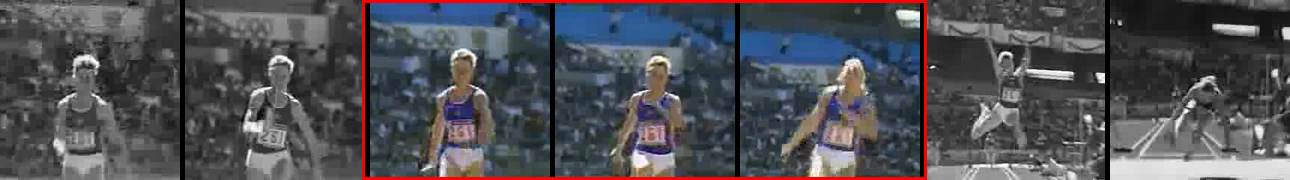}
        \vspace{-13pt}
        \caption{\textit{LongJump}}
        \label{fig:long_jump}
    \end{subfigure}
    \begin{subfigure}[t]{\linewidth}
        \centering
        \includegraphics[width=\linewidth]{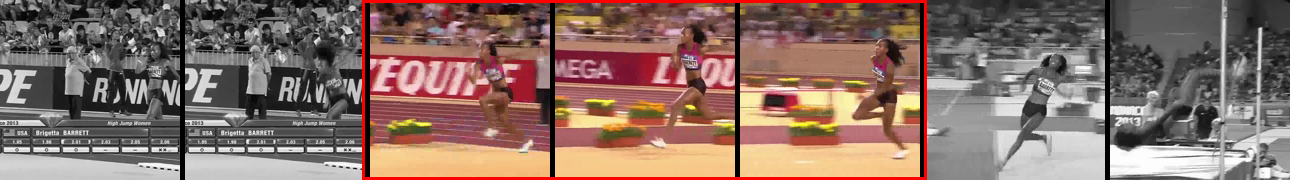}
        \vspace{-13pt}
        \caption{\textit{HighJump}}
        \label{fig:high_jump}
    \end{subfigure}
    
    \vspace{-10pt}
    \caption{\textbf{Challenges of center-focused sampling in action distinction.} Each sequence shows seven evenly sampled frames across the action duration, using examples from THUMOS14~\cite{THUMOS14}.}

    \vspace{-17pt}
    \label{fig:boundary}
\end{figure}

In the decoder, deformable cross-attention relies on sampling points near the center of reference points, typically determined by multiplying the reference width by 0.5.
However, this center-focused approach restricts the model from capturing the contextual information from the surrounding region, which is crucial for classifying action instances and determining their start and end boundaries.
Fig.~\ref{fig:boundary} shows the challenges of center-focused sampling through the example of actions like \textit{LongJump} and \textit{HighJump}, where both involve a similar running motion.
When the model focuses only on the central motion (red box), identifying these two actions is challenging because the surrounding context (gray frames), such as the final landing motion, provides essential cues for identifying them.
Furthermore, relying solely on the running motion makes it challenging to determine accurate start and end boundaries.
These observations suggest that the center-focused sampling strategy is insufficient for capturing the full context of action instances.

In this paper, we propose a multi-dilated gated encoder and a central-adjacent region integrated decoder for temporal action detection transformer (\modelname{}).
First, we introduce a multi-dilated gated encoder (\encodername{}), which replaces the previous multi-scale deformable attention and feedforward network in the encoder.
\encodername{} utilizes multi-dilated convolutions to capture diverse feature representations across multiple receptive fields, reducing redundant information in multi-scale features while preserving the benefits of utilizing multi-scale information.
Additionally, we present a central-adjacent region integrated decoder (\decodername{}), which combines both central- and adjacent-region sampling based on the deformable cross-attention mechanism.
By incorporating these two types of information, \decodername{} enables the detector to capture a complete contextual view for each detection query.
Extensive experiments demonstrate that \modelname{} achieves state-of-the-art performance on popular benchmarks, and our method is adaptable to existing query-based detection frameworks.

\noindent Our contributions are summarized as three-fold:
\begin{itemize}
    \item We propose \modelname{} that combines multi-dilated gated encoder and central-adjacent region integrated decoder to address the unique challenges of TAD.
    \item As shown in Fig.~\ref{fig:convergence}, our method consistently achieves faster convergence and improves performance when applied to existing query-based detectors.
    \item Our experiments demonstrate that our \modelname{} achieves state-of-the-art performance on THUMOS14, ActivityNet~v1.3, and HACS-Segment.
\end{itemize}

\section{Related Work}
\label{sec:related_work}

\noindent \textbf{Action Recognition}
Action recognition is a fundamental task in video analysis. It involves classifying video sequences into specific action categories.
I3D~\cite{I3D} extends the inception network by incorporating 3D convolutions, while R(2+1)D~\cite{R2+1D} improves efficiency by decomposing 3D convolutions into separate 2D spatial and 1D temporal operations.
TSP~\cite{TSP} introduces temporal channel shifting to model temporal dynamics effectively without adding computational overhead.
VideoMAEv2~\cite{VideoMAEv2} leverages masked reconstruction pretraining method based on transformer architecture for robust video representation learning.
InternVideo2~\cite{InternVideo2} leverages both large-scale training data and a highly scalable model, further enhancing video representation learning.
These models are utilized in various downstream tasks like TAD as a feature extraction method.

\noindent \textbf{Anchor-free Detector}
Anchor-free detectors~\cite{AFSD,TALLFormer,ActionFormer,Ti-FAD,ASL,TriDet,VideoMambaSuite,DyFADet} provide flexibility in localizing action instances by utilizing an asymmetric modeling approach.
ActionFormer~\cite{ActionFormer} improves TAD performance by leveraging a transformer-based architecture to capture long-range dependencies in video data.
TriDet~\cite{TriDet} utilizes the trident prediction scheme and its proposed architecture.
ActionMamba~\cite{VideoMambaSuite} improves the ActionFormer detector by utilizing Mamba~\cite{Mamba} architecture at the temporal feature extraction.
However, despite these advancements, anchor-free detectors still require hand-crafted components such as NMS to remove redundant proposals.

\noindent \textbf{Query-based Detector}
Query-based detectors, drawing inspiration from DETR~\cite{DETR}, employ a set-prediction mechanism that minimizes dependence on hand-crafted components, eliminating the necessity for NMS.
RTD-Net~\cite{RTD-Net} and ReAct~\cite{ReAct} utilize query-based detection approaches; however, they do not fully address one-to-one matching in their architectures.
TadTR~\cite{TadTR} introduces cross-window fusion, applying NMS only to overlapping areas in sliding windows, which partially reduces dependency on hand-crafted components.
DualDETR~\cite{DualDETR} divides the decoder into separate branches for instance-level and boundary-level decoding, whereas our \decodername{} does not address split branch but unifies comprehensive range information within a single decoder.
Both TadTR and DualDETR still rely on sliding windows, which require NMS to handle redundant areas, thereby limiting their applicability as a fully end-to-end detector.
In contrast, TE-TAD~\cite{TE-TAD} achieves a fully end-to-end approach for TAD by reformulating coordinate representation, removing the need for hand-crafted components like NMS and sliding windows.
Building on TE-TAD, we propose \modelname{}, which introduces \encodername{}, a multi-scale adapter, and \decodername{} for the decoder.
Although our \modelname{} is primarily based on TE-TAD, our method is designed to be adaptable across various query-based detectors by replacing the encoder with \encodername{} and decoder with \decodername{}.

\begin{figure*}[!t]
\centering
    \begin{tabular}{c}
        \includegraphics[width=0.94\linewidth]{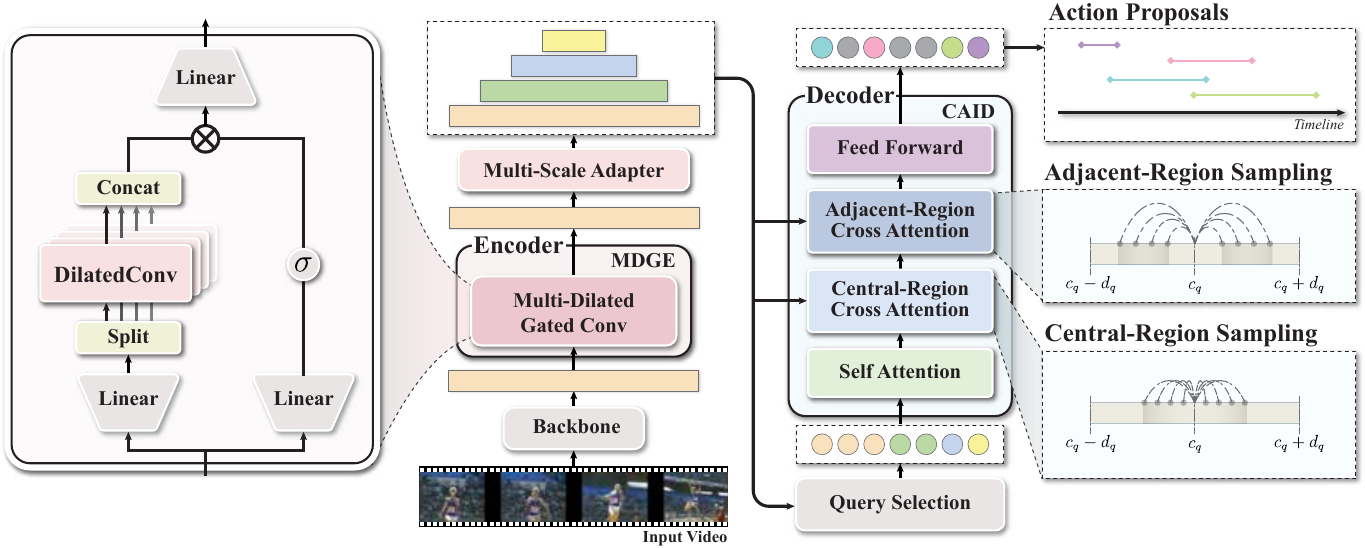}\\
    \end{tabular}
    \vspace{-9pt}
    \caption{\textbf{Overview of \modelname{}.} Our model processes video features through \encodername{} to capture distinct feature representations utilizing various receptive fields. Subsequently, \decodername{} captures both central and adjacent region information, enhancing action boundary regression and classification. For simplicity, residual connection and layer normalization are omitted.}
    \vspace{-15pt}
    \label{fig:model}
\end{figure*}

\section{Method}

\subsection{Preliminary}
\label{subsec:preliminary}
% \noindent \textbf{Temporal Action Detection}
Let $X \in \mathbb{R}^{T_0 \times C}$ represents the video feature sequence extracted by the backbone network, where $T_0$ denotes the temporal length of the sequence, and $C$ corresponds to the feature dimension.
Each element of the sequence, denoted as $X = \{ x_t \}_{t=1}^{T_0}$, is associated with a snippet at timestep $t$, with each snippet covering a few consecutive frames.
These snippets are processed using a pre-trained backbone network such as I3D~\cite{I3D} or InternVideo2~\cite{InternVideo2}.
Each video contains multiple action instances, each defined by start and end timestamps $s$ and $e$, as well as its action class $c$.
Formally, the set of action instances in a video is expressed as $\mathcal{A} = \{(s_n, e_n, c_n)\}_{n=1}^{N}$, where $N$ denotes the total number of action instances, and $s_n$, $e_n$, and $c_n$ represent the start time, end time, and action class of the $n$-th instance, respectively.
The primary objective of TAD is to predict the set of action instances $\mathcal{A}$ for a given video.

The query-based detectors~\cite{TadTR,DualDETR,TE-TAD} employ $N_q$ queries to detect action instances.
The set of queries is represented as $\mathcal{Q}^{(0)} = \{ F_q^{(0)},  (c_q^{(0)}, d_q^{(0)}) \}_{q=1}^{N_q}$, where $F_q^{(0)}$ is the initial embedding of the $q$-th query, $c_q^{(0)}$ and $d_q^{(0)}$ denote the initial center and width reference point of the $q$-th query, respectively.
These queries interact with the encoded features through deformable cross-attention layers in the decoder, where they are iteratively refined across each layer $l$:
\begin{equation}
    \mathcal{Q}^{(l)} = \text{Decoder}^{(l)}(\mathcal{Q}^{(l-1)}) \quad l = 1, \dots, L_D,
\end{equation}
where $L_D$ denotes the number of decoder layer.
The final refined queries $Q^{(L_D)}$ are then used for final predictions $\hat{\mathcal{A}}$ obtained through the classification and regression heads.

\subsection{\modelname{}}
In this part, we describe our method mainly based on TE-TAD, but our \modelname{} can be applied to various query-based detectors by simply replacing the previous encoder and decoder architecture.
The overall architecture of \modelname{} is illustrated in Fig.~\ref{fig:model}.
Our method mainly addresses three parts: (1) the multi-dilated gated encoder~(\encodername{}), (2) the multi-scale adapter, which converts the single-scale feature into multi-scale features, providing diverse scale information for query selection and decoder, and (3) the central-adjacent region integrated decoder~(\decodername{}).

\noindent \textbf{Embedding}
We project input features $X$ using a single convolutional neural network to align them with the dimension of the transformer architecture.
\begin{equation}
    \small Z^{(0)} = \textrm{LayerNorm}(\textrm{Conv}(X)),
\end{equation}
where $Z^{(0)} \in \mathbb{R}^{D \times T_0}$ represented the embedded features of the detector.
Here, $D$ denotes the width of the encoder and decoder.
In contrast to TE-TAD, we do not address the multi-scale features at the initial and encoding stages.

\newpage
\noindent \textbf{Multi-Dilated Gated Encoder (\encodername{})}
As discussed in Sec.~\ref{sec:intro}, both single-scale and multi-scale approaches have inherent limitations.
Single-scale methods struggle to capture long temporal dependencies.
Conversely, multi-scale methods can capture a broader range of temporal scales but cause highly correlated features across each scale.

To address the limitations of both approaches, we introduce \encodername{} that replaces the previous multi-scale deformable encoder structure.
Inspired by a previous work~\cite{DeepLab}, \encodername{} applies multi-dilated convolutions to capture diverse receptive fields within a single encoder.
This approach enables the model to extract short-term and long-term temporal features without relying on a multi-scale structure.
Furthermore, inspired by gating mechanism~\cite{GatedConv}, \encodername{} selectively filters out redundant information, retaining only the most relevant features across the different receptive fields.
In the following, we describe the detailed structure of \encodername{}, which is composed of $L_E$ multi-dilated gated convolution layers.

At each encoder layer $l$, the input features $Z^{(l-1)}$ from the previous layer are first projected into two paths:
\begin{equation}
    Z_{\text{conv}}^{(l)} = \text{Linear}(Z^{(l-1)})\text{,} \quad Z_{\text{gate}}^{(l)} = \text{Linear}(Z^{(l-1)})\text{,}
\end{equation}
where $Z_{\text{conv}}^{(l)} \in \mathbb{R}^{D_h \times T_0}$ and $Z_{\text{gate}}^{(l)} \in \mathbb{R}^{D_h \times T_0}$.
Here, $D_h$ represents the hidden dimension of the feedforward network within the transformer architecture.
Instead of utilizing the feedforward network in the encoder, we expand the feature dimension before applying the dilated convolution.
This approach retains a similar parameter to a single transformer layer that consists of an attention layer and a feedforward network.
Each transformed features, $Z_{\text{conv}}^{(l)}$ and $Z_{\text{gate}}^{(l)}$, are then processed in separate paths independently for multi-dilated convolution and gating mechanism, respectively.

The transformed features for multi-dilated convolution $Z_{\text{conv}}^{(l)}$ are split along the channel dimension into $N_d$ equal subsets, denoted as $Z_{\text{conv}}^{(l,i)} \in \mathbb{R}^{(D_h / N_d) \times T_0}$, where $N_d$ is the number of parallel dilated convolutions.
Each subset is processed by a dilated convolution with a different dilation rate, increasing from 1 up to $N_d$.
The output features for each subset are expressed as follows:
\begin{equation}
    Z_{\textrm{dilated}}^{(l,i)} = \text{DilatedConv}_{(l,d_i)}(Z_{\textrm{conv}}^{(l,i)}) \quad i = 1, \dots, N_d,
\end{equation}
where $d_i$ is the dilation rate for the $i$-th convolution, set as $d_i = i$.
The increasing dilation rates provide varying receptive field sizes, allowing the model to capture both short- and long-term temporal features simultaneously.
Our encoder has two main hyperparameters: the number of dilated convolutions $N_d$ and the kernel size.

Subsequently, the outputs of the dilated convolutions $Z_{\text{dilated}}^{(l,i)}$ are concatenated along the channel dimension:
\begin{equation}
    Z_{\text{concat}}^{(l)} = \textrm{Concat}(Z_{\text{dilated}}^{(l,1)}, \dots, Z_{\text{dilated}}^{(l,N_d)}).
\end{equation}

Following this concatenation, we apply a gating mechanism to selectively retain relevant features:
\begin{equation}
    Z^{(l)} = Z_{\text{concat}}^{(l)} \odot \sigma(Z_{\text{gate}}^{(l)}),
\end{equation}
where $\sigma$ is an activation function and $\odot$ denotes element-wise multiplication.
Specifically, we use the SiLU activation function for the gate activation.

Consequently, the encoding process across $L_E$ layers of \encodername{} can be summarized as follows:
\begin{equation}
    Z^{(l)} = \textrm{\encodername{}}^{(l)}(Z^{(l-1)}) \quad  l=1, \dots, L_E.
\end{equation}

Our \encodername{} design enables the encoder to capture diverse temporal relations by leveraging dilated convolutions and a gating mechanism without requiring explicit multi-scale features at each layer.

\newpage

\noindent \textbf{Multi-Scale Adapter \& Query Selection}
TE-TAD~\cite{TE-TAD} employs a two-stage approach with multi-scale features for query selection.
To take advantage of the two-stage approach and utilize multi-scale features for query selection, we introduce a multi-scale adapter that converts the single-scale feature to multi-scale features.
Our multi-scale adapter utilizes the encoder output feature $Z^{(L_E)}$ to generate multi-scale representations.
Specifically, $Z^{(L_E)}$ is progressively downsampled to produce a set of features at multiple levels, with each subsequent level reduced to half the temporal length of the previous one.
We denote the $L$ levels of multi-scale features as follows:
\begin{equation}
    F^{(l)} = \text{DownSample}(Z^{(L_E)}) + E^{(l)} \quad l = 1, \dots, L \text{,}
\end{equation}
where each $F^{(l)} \in \mathbb{R}^{T_l \times D}$ represents the resized feature at level $l$, with $T_l$ being half the length of the previous level.
To enable the decoder to distinguish between these multi-scale levels, we add a level-specific embedding $E^{(l)} \in \mathbb{R}^{1 \times D}$, applied consistently across all time steps.
These multi-scale features are then utilized for the query selection process.
Unlike the adaptive query selection (AQS) in TE-TAD~\cite{TE-TAD} that enforces a strict uniform sampling of queries across the video, we apply a top-$k$ selection approach based on binary classification scores from the encoder.

Subsequently, we utilize the top-$k$ indices to retrieve both the query embeddings and the corresponding reference points $(c_q, d_q)$ based on a time-aligned query generation method~\cite{TE-TAD} that assigns temporal center points $c_q$ and widths $d_q$, aligning them with their respective positions in the video.
Additionally, the query embedding for each selected query is obtained by linearly projecting the corresponding top-$k$ multi-scale features.
The input embedding of a decoder is denoted as follows:
\begin{equation}
    F_{q}^{(0)} = \text{LayerNorm}(\text{Linear}(F_{\text{topk}})) \text{,}
\end{equation}
where $F_{\text{topk}}$ denotes the top-$k$ selected features.

\noindent \textbf{Central-Adjacent Region Integrated Decoder (\decodername{})}  
The previous query-based methods~\cite{TadTR, TE-TAD} apply temporal deformable cross-attention that applies the center-focused sampling strategy.
As discussed in Sec. \ref{sec:intro}, the center-focused sampling does not provide sufficient information for detecting the action instances.
To address this issue, we introduce \decodername{}, which combines central- and adjacent-region cross-attention.
In standard deformable cross-attention, features are sampled around each reference point, defined by its center $c_q$ and duration $d_q$.
The sampling offset $\Delta p_{mqk}$ for each head $m$ and sampling point $k$ is computed as:
\begin{equation}
    \Delta p_{mqk} = \text{Linear}(F_q) = W F_q + b \text{,}
    \label{eq:sampling_offset}
\end{equation}
where $W$ and $b$ are learnable parameters for obtaining the sampling offset by linear projection.
Generally, $W$ is initialized to zero and $b$ within the range $[-1, 1]$.
Using the computed offset $\Delta p_{mqk}$, sampling points $p_{mqk}$ are determined based on the reference points as:
\begin{equation}
    p_{mqk} = c_q + 0.5 w_q \Delta p_{mqk} \text{,}
    \label{eq:sampling_point}
\end{equation}
where the factor $0.5 w_q$ ensures the sampling points are initially positioned close to the center of each reference point.
In our approach, we do not change how to obtain the sampling points when addressing central- and adjacent-region sampling.
We change the initialization method for the bias value $b$, which determines the initial sampling points.

As shown in Fig.~\ref{fig:model}, our \decodername{} contains two cross-attention layers sequentially: central-region cross-attention and adjacent-region cross-attention.
For central-region cross-attention, the initial bias of sampling offsets $b$ are initialized uniformly within the range $[-1, 1]$, which is identical to the previous methods~\cite{TadTR,TE-TAD,Deformable-DETR,DAB-DETR,DN-DETR,DINO}.
As Eq.~\eqref{eq:sampling_point}, this initialization constrains the initial bias for sampling points $b$ within $[-0.5, 0.5]$, focusing on the central region of the reference points.
For adjacent-region cross-attention, the bias of sampling offsets $b$ are adjusted to capture surrounded points of central-region cross-attention.
Specifically, half of the sampling offsets are initialized within the range $[-1.5, -0.5]$ to focus on the left, while the other half are initialized within the range $[0.5, 1.5]$ to focus on the right.
This adjustment, combined with Eq.~\eqref{eq:sampling_point}, constraints in the initial sampling points $p_{mqk}$ within $[-0.75, -0.25]$ and $[0.25, 0.75]$.
Overall, we sequentially apply self-attention, central-region cross-attention, adjacent-region cross-attention, and feedforward network across all $L_D$ layers, as shown in Fig. \ref{fig:model}.

\begin{table*}[!ht]
\centering

\resizebox{0.93\linewidth}{!}{
\begin{tabular}{cccccccccccccccc}
\Xhline{2\arrayrulewidth}
\noalign{\smallskip}
\multirow{2.7}{*}{\textbf{\shortstack{Training\\Type}}} & \multirow{2.7}{*}{\textbf{\shortstack{Head\\Type}}} & \multirow{2.7}{*}{\textbf{Method}} & \multirow{2.7}{*}{\textbf{Feature}} & & \multicolumn{6}{c}{\textbf{THUMOS14}} & & \multicolumn{4}{c}{\textbf{ActivityNet v1.3}}\\
\noalign{\smallskip}
\cline{6-11}
\cline{13-16}
\noalign{\smallskip}
& & & & & 0.3 & 0.4 & 0.5 & 0.6 & 0.7 & Avg. & & 0.5 & 0.75 & 0.95 & Avg.\\
\noalign{\smallskip}
\Xhline{2\arrayrulewidth}
\noalign{\smallskip}

\multirow{4}{*}{Full} &
\multirow{4}{*}{-} &

AFSD~\cite{AFSD} & I3D~\cite{I3D} & &
67.3 & 62.4 & 55.5 & 43.7 & 31.1 & 52.0 & & 52.4 & 35.3 & 6.5 & 34.4\\
& & TALLFormer~\cite{TALLFormer} & Swin-B~\cite{VideoSwin} & &
76.0 & - & 63.2 & - & 34.5 & 59.2 & & 54.1 & 36.2 & 7.9 & 35.6\\
& & ViT-TAD~\cite{ViT-TAD} & ViT-B~\cite{ViT} & &
85.1 & 80.9 & 74.2 & 61.8 & 45.4 & 69.5 & & 55.8 & 38.5 & 8.8 & 37.4\\
& & AdaTAD~\cite{AdaTAD} & VideoMAEv2-g~\cite{VideoMAEv2} & &
\textbf{89.7} & \textbf{86.7} & \textbf{80.9} & \textbf{71.0} & \textbf{56.1} & \textbf{76.9} & & \textbf{61.7} & \textbf{43.4} & \textbf{10.9} & \textbf{41.9}\\

\noalign{\smallskip}
\hline
\noalign{\smallskip}

\multirow{19}{*}{\shortstack{Head\\only}} &
\multirow{5}{*}{\shortstack{Anchor\\-free}} &
TriDet~\cite{TriDet} & I3D~\cite{I3D} / R(2+1)D~\cite{R2+1D} & &
83.6 & 80.1 & 72.9 & 62.4 & 47.4 & 69.3 & & 54.7 & 38.0 & 8.4 & 36.8\\
& & DyFADet~\cite{DyFADet} & I3D~\cite{I3D} / R(2+1)D~\cite{R2+1D} & & 84.0 & 80.1 & 72.7 & 61.1 & 47.9 & 69.2 & & 58.1 & 39.6 & 8.4 & 38.5\\
& & ActionFormer~\cite{ActionFormer} & I3D~\cite{I3D} / R(2+1)D~\cite{R2+1D} & & 
82.1 & 77.8 & 71.0 & 59.4 & 43.9 & 66.8 & & 54.7 & 37.8 & 8.4 & 36.6\\
& & ActionFormer~\cite{ActionFormer} & InternVideo2~\cite{InternVideo2} & &  82.3 & 81.9 & 75.1 & 65.8 & 50.3 & 71.9 & & 61.5 & \textbf{44.6} & \textbf{12.7} & 41.2\\
& & ActionMamba~\cite{VideoMambaSuite} & InternVideo2~\cite{InternVideo2} & &
\textbf{86.9} & \textbf{83.1} & \textbf{76.9} & \textbf{65.9} & \textbf{50.8} & \textbf{72.7} & & \textbf{62.4} & 43.5 & 10.2 & \textbf{42.0}\\

\noalign{\smallskip}
\cline{2-16}
\noalign{\smallskip}

& \multirow{13.5}{*}{\shortstack{Query\\-based}}
& RTD-Net~\cite{RTD-Net} & I3D~\cite{I3D} / TSN~\cite{TSN} & &
68.3 & 62.3 & 51.9 & 38.8 & 23.7 & 49.0 & & 47.2 & 30.7 & 8.6 & 30.8\\
& & ReAct~\cite{ReAct} & I3D~\cite{I3D} / TSN~\cite{TSN} & &
69.2 & 65.0 & 57.1 & 47.8 & 35.6 & 55.0 & & 49.6 & 33.0 & 8.6 & 32.6\\
& & Self-DETR~\cite{Self-DETR} & I3D~\cite{I3D} & &
74.6 & 69.5 & 60.0 & 47.6 & 31.8 & 56.7 & & 52.3 & 33.7 & 8.4 & 33.8\\
& & TadTR~\cite{TadTR} & I3D~\cite{I3D} / R(2+1)D~\cite{R2+1D} & &
74.8 & 69.1 & 60.1 & 46.6 & 32.8 & 56.7 & & 53.6 & 37.5 & 10.6 & 36.8\\
& & DualDETR~\cite{DualDETR} & I3D~\cite{I3D} & & 82.9 & 78.0 & 70.4 & 58.5 & 44.4 & 66.8 & & 52.6 & 35.0 & 7.8 & 34.3\\
& & TE-TAD~\cite{TE-TAD} & I3D~\cite{I3D} / R(2+1)D~\cite{R2+1D} & &
83.3 & 78.4 & 71.3 & 60.7 & 45.6 & 67.9 & & 54.2 & 38.1 & 10.6 & 37.1\\
& & \baseline{\modelname{}$^\dagger$} & \baseline{I3D~\cite{I3D} / R(2+1)D~\cite{R2+1D}} & \baseline{} &
\baseline{81.6} & \baseline{77.7} & \baseline{70.3} & \baseline{60.5} & \baseline{48.4} & \baseline{67.7} & \baseline{} & \baseline{54.3} & \baseline{38.4} & \baseline{10.6} & \baseline{37.2}\\
& & \baseline{\modelname{}} & \baseline{I3D~\cite{I3D} / R(2+1)D~\cite{R2+1D}} & \baseline{} &
\baseline{83.6} & \baseline{79.6} & \baseline{71.9} & \baseline{61.5} & \baseline{48.6} & \baseline{69.0} & \baseline{} & \baseline{54.4} & \baseline{38.2} & \baseline{10.7} & \baseline{37.3}\\

\noalign{\smallskip}
\cdashline{3-16}
\noalign{\smallskip}
& & TadTR~\cite{TadTR} & InternVideo2~\cite{InternVideo2} & &
84.8 & 79.3 & 70.4 & 58.2 & 43.8 & 67.3 & & 57.1 & 38.8 & 11.0 & 38.2 \\ 

& & \baseline{TadTR~\cite{TadTR} + Ours} & \baseline{InternVideo2~\cite{InternVideo2}} & \baseline{} &
\baseline{86.1} & \baseline{81.8} & \baseline{73.7} & \baseline{61.7} &  \baseline{46.3} & \baseline{69.9} & \baseline{} & \baseline{60.2} & \baseline{41.0} & \baseline{11.2} & \baseline{40.5}\\ 
& & TE-TAD~\cite{TE-TAD} & InternVideo2~\cite{InternVideo2} & &
84.3 & 81.1 & 73.7 & 62.6 & 49.5 & 70.3 & & 61.3 & 41.8 & 10.9 & 41.1\\

& & \baseline{\modelname{}$^\dagger$} & \baseline{InternVideo2~\cite{InternVideo2}} & \baseline{} & \baseline{85.7} & \baseline{82.3} & \baseline{75.6} & \baseline{65.6} & \baseline{51.4} & \baseline{72.1} & \baseline{} & \baseline{58.9} & \baseline{43.4} & \baseline{11.4} & \baseline{41.3}\\
& & \baseline{\modelname{}}& \baseline{InternVideo2~\cite{InternVideo2}} & \baseline{} &
\baseline{\textbf{87.6}} & \baseline{\textbf{84.2}} & \baseline{\textbf{77.6}} & \baseline{\textbf{67.3}} & \baseline{\textbf{52.5}} & \baseline{\textbf{73.8}} & \baseline{} & \baseline{\textbf{62.0}} & \baseline{\textbf{43.1}} & \baseline{\textbf{11.3}} & \baseline{\textbf{42.0}}\\

\noalign{\smallskip}
\Xhline{2\arrayrulewidth}

\end{tabular}
}
\vspace{-7pt}
\caption{\textbf{Performance comparison with state-of-the-art methods on THUMOS14 and ActivityNet v1.3.}  In cases marked with $\dagger$, our method does not utilize  NMS. For TadTR + Ours, we additionally apply our \encodername{} and \decodername{} on TadTR.}

\vspace{-11pt}
\label{table:thumos}
\end{table*}

\subsection{Training and Inference}
\label{sec:training_and_inference}
\noindent \textbf{Training}
Following the previous works~\cite{TadTR,TE-TAD}, we use the bipartite matching loss~\cite{DETR}.
The total loss $\mathcal{L}_{total}$ is defined as follows:
\begin{equation}
    \mathcal{L}_{total}(\mathcal{A}, \hat{\mathcal{A}}) = \sum_{i=1}^{N_q}{\mathcal{L}_{match} (\mathcal{A}_i, \hat{\mathcal{A}}_{\pi(i)})},
    \label{eq:total_loss}
\end{equation}
where $\mathcal{L}_{match}$ denotes the bipartite matching loss, which considers both classification and regression loss between ground truth $\mathcal{A}_i$ and predicted action instances $\hat{\mathcal{A}}_{\pi(i)}$ from the last layer of the decoder.
The permutation indices $\pi(i)$ are obtained through bipartite matching~\cite{HungarianMatching}.
This cost function $\mathcal{L}_{match}$ is a composite of the classification, and regression loss.
We use focal loss~\cite{FocalLoss} for the classification loss to effectively manage class imbalance.
For the regression loss, our method incorporates GIoU~\cite{GIoU} and a log-ratio distance loss~\cite{TE-TAD}.

\noindent \textbf{Inference}
Following the previous work~\cite{TE-TAD} that removes the need for post-processing steps, the predictions of \modelname{} from the final layer of the decoder $\hat{\mathcal{A}}$ are directly used.
For a fair comparison, we report both raw prediction results and NMS applied results.

\section{Experiments}
\label{sec:experiments}
\subsection{Setup}
\noindent \textbf{Datasets} We conduct experiments on three datasets: THUMOS14~\cite{THUMOS14}, ActivityNet~v1.3~\cite{ActivityNet}, and HACS-Segment~\cite{HACS}.
THUMOS14 consists of 20 action classes with 200 validation and 213 test videos, containing 3,007 and 3,358 action instances, respectively.
ActivityNet~v1.3 is a large-scale dataset with 200 action classes, including 10,024 videos for training and 4,926 videos for validation.
HACS-Segment is another large-scale TAD dataset with extensive annotations, covering 200 activity classes similar to ActivityNet~v1.3.
It provides 37,613 videos for training and 5,981 videos for validation.
These datasets provide a rigorous evaluation environment for our method, containing diverse actions and scenes.

\noindent \textbf{Evaluation Metric}
We follow the standard evaluation protocol for all datasets, utilizing mAP at different intersections over union (IoU) thresholds to evaluate TAD performance.
The IoU thresholds for THUMOS14 are set at [0.3:0.7:0.1], while for ActivityNet~v1.3 and HACS-Segment, the results are reported at IoU threshold [0.5, 0.75, 0.95] with the average mAP computed at [0.5:0.95:0.05].

\noindent \textbf{Implementation Details}
We describe the implementation details for each dataset in the Supplementary Sec. \ref{sec:implementation_details}.

\subsection{Main Results}

\noindent \textbf{THUMOS14}
Table~\ref{table:thumos} contains a comparison with the state-of-the-art methods on THUMOS14.
Our \modelname{} shows consistent improvements over TadTR~\cite{TadTR} and TE-TAD~\cite{TE-TAD} on both I3D~\cite{I3D} and InternVideo2~\cite{InternVideo2} features.
Even without applying NMS, our method outperforms the existing query-based detectors. 
Furthermore, our model outperforms the existing snippet-based head-only training methods, and our \modelname{} shows a comparable performance even compared to the full training method.

\noindent \textbf{ActivityNet~v1.3}
Following the conventional approach~\cite{ActionFormer,TriDet,TadTR,TE-TAD}, the external classification score is used to evaluate ActivityNet~v1.3.
The pre-extracted classification scores are combined with class-agnostic predictions from a binary detector to obtain class labels.
For R(2+1)D~\cite{R2+1D} and InternVideo2~\cite{InternVideo2} features, classification results from CUHK~\cite{CUHK} and InternVideo2~\cite{InternVideo2} are incorporated to obtain class scores, respectively.
As demonstrated in Table~\ref{table:thumos}, \modelname{} achieves consistent improvements over TadTR and TE-TAD on ActivityNet~v1.3.
Furthermore, \modelname{} demonstrates competitive performance compared to other types of state-of-the-art methods, demonstrating effectiveness across various datasets and feature extractors.

\begin{table}[!t]
\centering
\resizebox{\linewidth}{!}{
\begin{tabular}{cccccccc}
\Xhline{2\arrayrulewidth}
\noalign{\smallskip}
\multirow{2.7}{*}{\textbf{\shortstack{Head\\Type}}} & \multirow{2.7}{*}{\textbf{Method}} & \multirow{2.7}{*}{\textbf{Feature}} & \multicolumn{4}{c}{\textbf{mAP}}\\
\noalign{\smallskip}
\cline{4-7}
\noalign{\smallskip}
& & & 0.5 & 0.75 & 0.95 & Avg.\\
\noalign{\smallskip}
\Xhline{2\arrayrulewidth}
\noalign{\smallskip}

\multirow{4}{*}{\shortstack{Anchor\\-based}}
& SSN~\cite{SSN} &  I3D~\cite{I3D} &
28.8 & 18.8 & 5.3 & 19.0\\
& G-TAD~\cite{G-TAD} & I3D~\cite{I3D} &
41.1 & 27.6 & 8.3 & 27.5\\
& BMN~\cite{BMN} & SlowFast~\cite{SlowFast} &
52.5 & 36.4 & 10.4 & 35.8\\
& TCANet~\cite{TCANet} & SlowFast~\cite{SlowFast} &
54.1 & 37.2 & 11.3 & 36.8\\
\noalign{\smallskip}
\hline
\noalign{\smallskip}

\multirow{8}{*}{\shortstack{Anchor\\-free}}
& TALLFormer~\cite{TALLFormer} & Swin-B~\cite{VideoSwin}& 
55.0 & 36.1 & 11.8 & 36.5\\
& TriDet~\cite{TriDet} & I3D~\cite{I3D} &
54.5 & 36.8 & 11.5 & 36.8\\
& TriDet~\cite{TriDet} & SlowFast~\cite{SlowFast} &
56.7 & 39.3 & 11.7 & 38.6\\
& TriDet~\cite{TriDet} & VideoMAEv2-g~\cite{VideoMAEv2} &
62.4 & 44.1 & 13.1 & 43.1\\
& DyFADet~\cite{DyFADet} & SlowFast~\cite{SlowFast} & 
57.8 & 39.8 & 11.8 & 39.2\\
& DyFADet~\cite{DyFADet} & VideoMAEv2-g~\cite{VideoMAEv2} & 
64.0 & 44.8 & 14.1 & 44.3\\
& ActionFormer~\cite{ActionFormer} & InternVideo2~\cite{InternVideo2} & 62.6 & 44.6 & 12.7 & 43.3\\
& ActionMamba~\cite{VideoMambaSuite} & InternVideo2~\cite{InternVideo2} & \textbf{64.0} & \textbf{45.7} & \textbf{13.3} & \textbf{44.6}\\

\noalign{\smallskip}
\hline
\noalign{\smallskip}

\multirow{6}{*}{\shortstack{Query\\-based}}
& TadTR~\cite{TadTR} & I3D~\cite{I3D} &
47.1 & 32.1 & 10.9 & 32.1\\

& TadTR~\cite{TadTR} & InternVideo2~\cite{InternVideo2} & 54.2 & 38.8 & 12.8 & 37.8 \\ 
& \baseline{TadTR~\cite{TadTR} + Ours} & \baseline{InternVideo2~\cite{InternVideo2}} & \baseline{55.0} & \baseline{40.0} & \baseline{13.8} & \baseline{39.0}\\

& TE-TAD~\cite{TE-TAD} & InternVideo2~\cite{InternVideo2} & 60.4 & 45.6 & 16.5 & 44.1 \\
& \baseline{\modelname{}$^\dagger$} & \baseline{InternVideo2~\cite{InternVideo2}} &
\baseline{61.1} & \baseline{47.5} & \baseline{\textbf{17.8}} & \baseline{45.5}\\

& \baseline{\modelname{}} & \baseline{InternVideo2~\cite{InternVideo2}} &
\baseline{\textbf{62.4}} & \baseline{\textbf{47.9}} & \baseline{17.6} & \baseline{\textbf{45.9}}\\

\noalign{\smallskip}
\Xhline{2\arrayrulewidth}
\noalign{\smallskip}

\end{tabular}
}
\vspace{-10pt}
\caption{\textbf{Performance comparison with state-of-the-art methods on HACS-Segment.} In cases marked with $\dagger$, our method does not utilize NMS. For TadTR + Ours, we additionally apply our \encodername{} and \decodername{} on TadTR.}
\vspace{-12pt}
\label{table:hacs}
\end{table}
\noindent \textbf{HACS-Segment}
Table~\ref{table:hacs} presents a comparison of our model with state-of-the-art methods on HACS-Segment.
Our model achieves significant improvements over previous methods, establishing a new state-of-the-art performance, including mAP at higher IoU thresholds of 0.75 and 0.95.
The stronger performance at the 0.95 IoU threshold indicates that \modelname{} excels in precise action localization.
Moreover, the superior performance of \modelname{} on this larger dataset, compared to THUMOS14 and ActivityNet~v1.3, demonstrates its scalability and robustness across varying data sizes and complexities.

\begin{figure}[!h]
    \centering
    \includegraphics[width=0.62\linewidth]{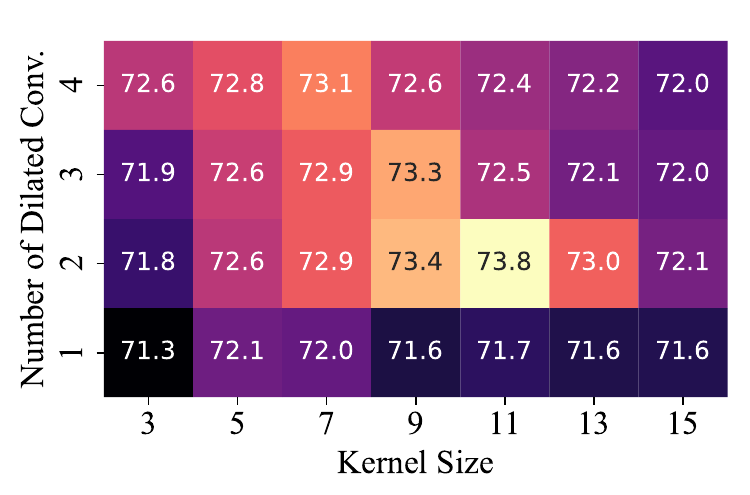}\\ 
    \vspace{-10pt}
    \caption{\textbf{Ablation study on MDGE with InternVideo2 features on THUMOS14.} The heatmap shows mAP values for different combinations of kernel size and number of dilated convolutions.}

    \vspace{-15pt}
    \label{fig:enc_ablation}
\end{figure}
\subsection{Further Analysis}

\noindent \textbf{Ablation Study on \encodername{}}
Fig.~\ref{fig:enc_ablation} shows an ablation study on hyperparameters of \encodername{}, specifically the impact of kernel size and the number of dilated convolution layers $N_d$ on mAP performance.
Excluding the configuration with a single dilated convolution that consistently yields lower mAP values, configurations with multiple dilated layers demonstrate improved performance.
These results indicate that addressing diverse scale information is crucial for the encoder by utilizing our multi-dilated convolution that enables the model to capture diverse temporal relations.
Furthermore, compared to the TE-TAD~\cite{TE-TAD} baseline of 70.4 in Table \ref{table:contribution}, our \encodername{} consistently improves performance regardless of hyperparameter variations, demonstrating robustness across different configurations.

\begin{table}[!t]
    \centering
    \resizebox{0.70\linewidth}{!}{
        \begin{tabular}{ccc}
            \Xhline{2\arrayrulewidth}
            \noalign{\smallskip}
            & \textbf{Decoder Sequece} & \textbf{mAP@AVG}\\
            \noalign{\smallskip}
            \Xhline{2\arrayrulewidth}
            \noalign{\smallskip}
            \#1 & SA $\rightarrow$ CCA $\rightarrow$ FFN &  72.5\\
            \#2 & SA $\rightarrow$ CCA (1.5$\times$) $\rightarrow$ FFN &  72.6\\
            \#3 &SA $\rightarrow$ CCA (2.0$\times$) $\rightarrow$ FFN &  71.8\\

            \noalign{\smallskip}
            \hline
            \noalign{\smallskip}

            \#4 & SA $\rightarrow$ ACA $\rightarrow$ FFN &  69.2\\

            \noalign{\smallskip}
            \hline
            \noalign{\smallskip}
            
            \#5 & SA $\rightarrow$ CCA $\rightarrow$ CCA $\rightarrow$ FFN & 72.7\\
            \#6 & SA $\rightarrow$ CCA $\rightarrow$ ACA $\rightarrow$ FFN & \textbf{73.8}\\
            \#7 & SA $\rightarrow$ ACA $\rightarrow$ CCA $\rightarrow$ FFN & 73.2\\
            \noalign{\smallskip}
            \Xhline{2\arrayrulewidth}
        \end{tabular}
    }
    \vspace{-7pt}
    \caption{\textbf{Analysis of decoder operation sequences using InternVideo2 features on THUMOS14.} SA refers to the self-attention layer. CCA denotes central-region cross-attention (as used in previous works such as~\cite{TadTR, TE-TAD}). CCA (1.5$\times$ and 2.0$\times$) indicates an expansion of the initial sampling range of central-region cross-attention by 1.5 and 2.0 times, respectively, to cover a broader area around the reference points. ACA refers to adjacent-region cross-attention, and FFN represents the feedforward network.}

    \vspace{-7pt}
    \label{table:decoder}
\end{table}

\noindent \textbf{Ablation Study on \decodername{}}
Table \ref{table:decoder} demonstrates the effects of integrating the proposed adjacent-region cross-attention (ACA) into the decoder operation sequence.
We evaluate several configurations, including variations of central-region cross-attention (CCA) with expanded initialization ranges (1.5$\times$ and 2.0$\times$).
CCA (1.5$\times$) and CCA (2.0$\times$) (Rows \#2 and \#3) show that simply increasing the initialization range of CCA to cover a broader area, similar range to \decodername{}, does not show the improved performance compared to the original setting (Row \#1).
This indicates that simply expanding the initialization range does not benefit the detector.
Moreover, when using ACA alone (Row \#4), we observe a significant decrease in performance, underscoring the importance of balancing adjacent-region information with central-region information.
Configurations that use two successive CCA layers (Row \#5) show only marginal improvements, suggesting that additional central-region cross-attention alone does not significantly enhance performance.
The highest mAP score is achieved by the sequence of SA $\rightarrow$ CCA $\rightarrow$ ACA $\rightarrow$ FFN, which combines central-region cross-attention and adjacent-region cross-attention, which is our \decodername{} (Row \#6).

\begin{table}[!t]
    \centering
    \resizebox{0.97\linewidth}{!}{
        \begin{tabular}{c  cc  cc  cccc}
            \Xhline{2\arrayrulewidth}
            \noalign{\smallskip}
            \multirow{2.5}{*}{\textbf{Baseline}} & \multirow{2.5}{*}{\textbf{Enc.}} & \multirow{2.5}{*}{\textbf{Dec.}} &  \multirow{2.5}{*}{\textbf{\shortstack{\encodername{}}}} & \multirow{2.5}{*}{\textbf{\decodername{}}} & \multicolumn{4}{c}{\textbf{mAP}}\\
            \noalign{\smallskip}
            \cline{6-9}
            \noalign{\smallskip}
            & & & & & 0.3 & 0.5 & 0.7 & Avg.\\
            \noalign{\smallskip}
            \Xhline{2\arrayrulewidth}
            \noalign{\smallskip}
            \multirow{4}{*}{\shortstack{TadTR~\cite{TadTR}}}
            & \multirow{4}{*}{S} & \multirow{4}{*}{S}& & & 84.8 & 70.4 & 43.8 & 67.3\\
            & & & \cmark & & 84.9 & 71.9 & 44.7 & 68.4\\
            & & & & \cmark & 84.4 & 70.8 & 45.4 & 67.9\\
            & & & \cmark & \cmark & 86.1 & 73.7 & 46.3 & 69.9\\
            \noalign{\smallskip}
            \hline
            \noalign{\smallskip}
            \multirow{6.5}{*}{\shortstack{TE-TAD~\cite{TE-TAD}}}
            & \multirow{2}{*}{M} & \multirow{2}{*}{M} & & & 84.3 & 73.7 & 49.5 & 70.3\\
            &  & & & \cmark & 85.2 & 74.6 & 49.9 & 70.8\\
            \noalign{\smallskip}
            \cline{2-9}
            \noalign{\smallskip}
            & \multirow{4}{*}{S} & \multirow{4}{*}{M} & & & 85.4 & 73.8 & 49.1 & 70.4\\
            & & & \cmark & & 87.0 & 75.8 & 51.9 & 72.5\\
            & & & & \cmark & 85.5 & 74.6 & 51.0 & 71.4\\
            & & & \cmark & \cmark & \textbf{87.6} & \textbf{77.6} & \textbf{52.5} & \textbf{73.8}\\
            \noalign{\smallskip}
            \Xhline{2\arrayrulewidth}
        \end{tabular}
    }
    \vspace{-7pt}
    \caption{\textbf{Ablation study on the contributions of each component using InternVideo2 features on THUMOS14.} The first row for TadTR~\cite{TadTR} and TE-TAD~\cite{TE-TAD} represents the baseline performance. Enc. and Dec. refer to how scale information is handled in the encoder and decoder. S: single-scale. M: multi-scale.}
    \vspace{-12pt}
    \label{table:contribution}
\end{table}

\noindent \textbf{Component Contribution Analysis}  
Table~\ref{table:contribution} provides an analysis of the performance contributions from each proposed component, evaluated using InternVideo2~\cite{InternVideo2} features on THUMOS14~\cite{THUMOS14}.
The first row for each method shows the baseline performance without our proposed enhancements, allowing for a direct comparison with subsequent configurations.
Notably, converting multi-scale encoding to single-scale encoding alone in TE-TAD~\cite{TE-TAD} results in minimal performance change (70.3 $\rightarrow$ 70.4).
This result aligns with our motivation, as shown in Fig. \ref{fig:cka}, which is that the previous multi-scale features contain redundant information.
Furthermore, both \encodername{} and \decodername{} show consistent improvements in both TadTR and TE-TAD, demonstrating the robustness and effectiveness of our approach across different baseline architectures.

\begin{table}[!t]
    \centering
    \resizebox{0.75\linewidth}{!}{
        \begin{tabular}{c  c  cccc}
            \Xhline{2\arrayrulewidth}
            \noalign{\smallskip}
            \multirow{2.5}{*}{\textbf{Baseline}} & \multirow{2.5}{*}{\textbf{AQS~\cite{TE-TAD}}} & \multicolumn{4}{c}{\textbf{mAP}}\\
            \noalign{\smallskip}
            \cline{3-6}
            \noalign{\smallskip}
            & & 0.3 & 0.5 & 0.7 & Avg.\\
            \noalign{\smallskip}
            \Xhline{2\arrayrulewidth}
            \noalign{\smallskip}
            \multirow{2}{*}{\shortstack{TE-TAD~\cite{TE-TAD}}} & \cmark & 85.6 & 73.7 & 47.6 & 70.3\\
            & & 82.5 & 71.8 & 48.4 & 68.7\\
            \noalign{\smallskip}
            \hline
            \noalign{\smallskip}
            \multirow{2}{*}{\shortstack{\modelname{}}}
            & \cmark & 85.8 & 76.4 & 51.5 & 72.2\\
            % 85.81 81.85 76.39 65.55 51.53 72.23
            & & \textbf{87.6} & \textbf{77.6} & \textbf{52.5} & \textbf{73.8}\\
            \noalign{\smallskip}
            \Xhline{2\arrayrulewidth}
        \end{tabular}
    }
    \vspace{-7pt}
    \caption{\textbf{Ablation study on the query selection method using InternVideo2 features on THUMOS14.}}
    \vspace{-9pt}
    \label{table:aqs}
\end{table}

\begin{figure}[!t]
    \centering
    \includegraphics[width=0.85\linewidth]{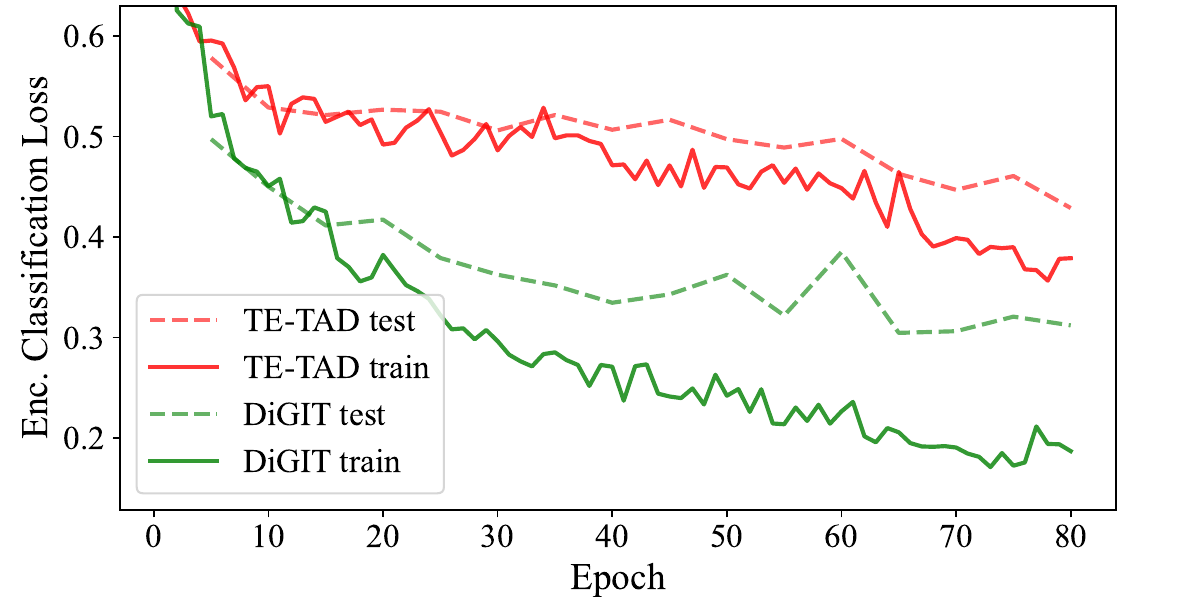}\\ 
    \vspace{-10pt}
    \caption{\textbf{Comparison of training and testing loss for the encoder classification matching loss using InternVideo2 features on THUMOS14.} Our method significantly accelerates the convergence of the encoder.}
    \vspace{-14pt}
    \label{fig:enc_loss}
\end{figure}

\noindent \textbf{Effect of \encodername{}}
In TE-TAD~\cite{TE-TAD}, adaptive query selection (AQS) is employed to sample initial queries across the entire video sequence uniformly.
This process enforces a strict uniform selection of queries to prevent queries from being overly concentrated in certain areas, as might happen with a simple top-k selection.
However, our observations suggest that this strict uniform condition is unnecessary when using a well-trained encoder, as provided by our MDGE.
As shown in Table 5, removing AQS results in higher mAP scores for DiGIT, whereas the performance of TE-TAD declines without AQS.
Furthermore, as illustrated in Fig. 6, DiGIT exhibits faster convergence and consistently lower loss values.
These results indicate that our MDGE enhances the representational ability of the encoder and can reduce the heuristic part of the query selection, which positively impactsoverall detection performance.

\subsection{Qualitative Results}

\begin{figure}[!t]
    \centering
    \hfill
    \begin{subfigure}[t]{0.48\linewidth}
        \centering
        \includegraphics[width=\linewidth]{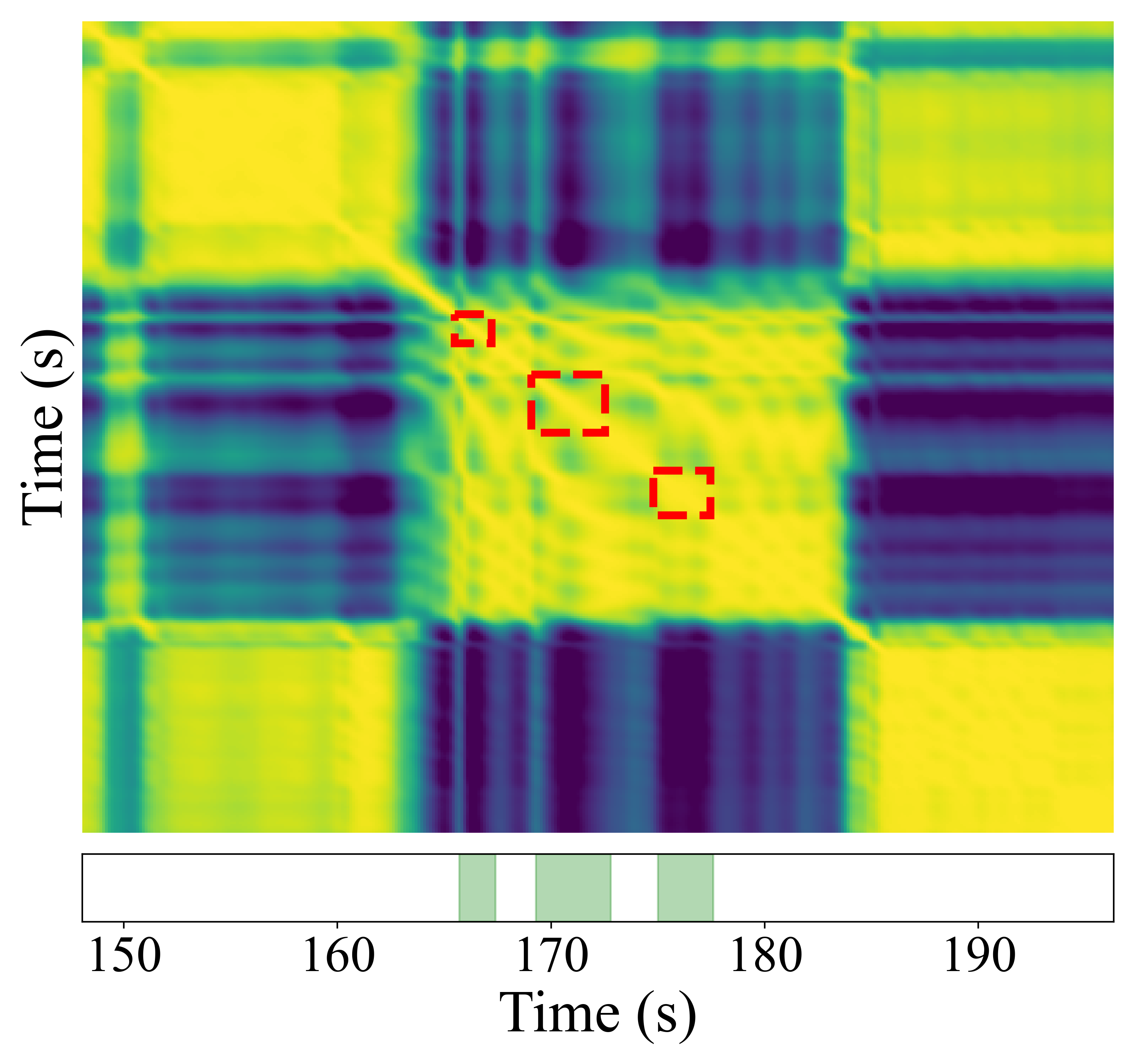}
        \vspace{-15pt}
        \caption{TE-TAD~\cite{TE-TAD}}
        \label{fig:cka_dino}
    \end{subfigure}%
    \hfill
    \begin{subfigure}[t]{0.48\linewidth}
        \centering
        \includegraphics[width=\linewidth]{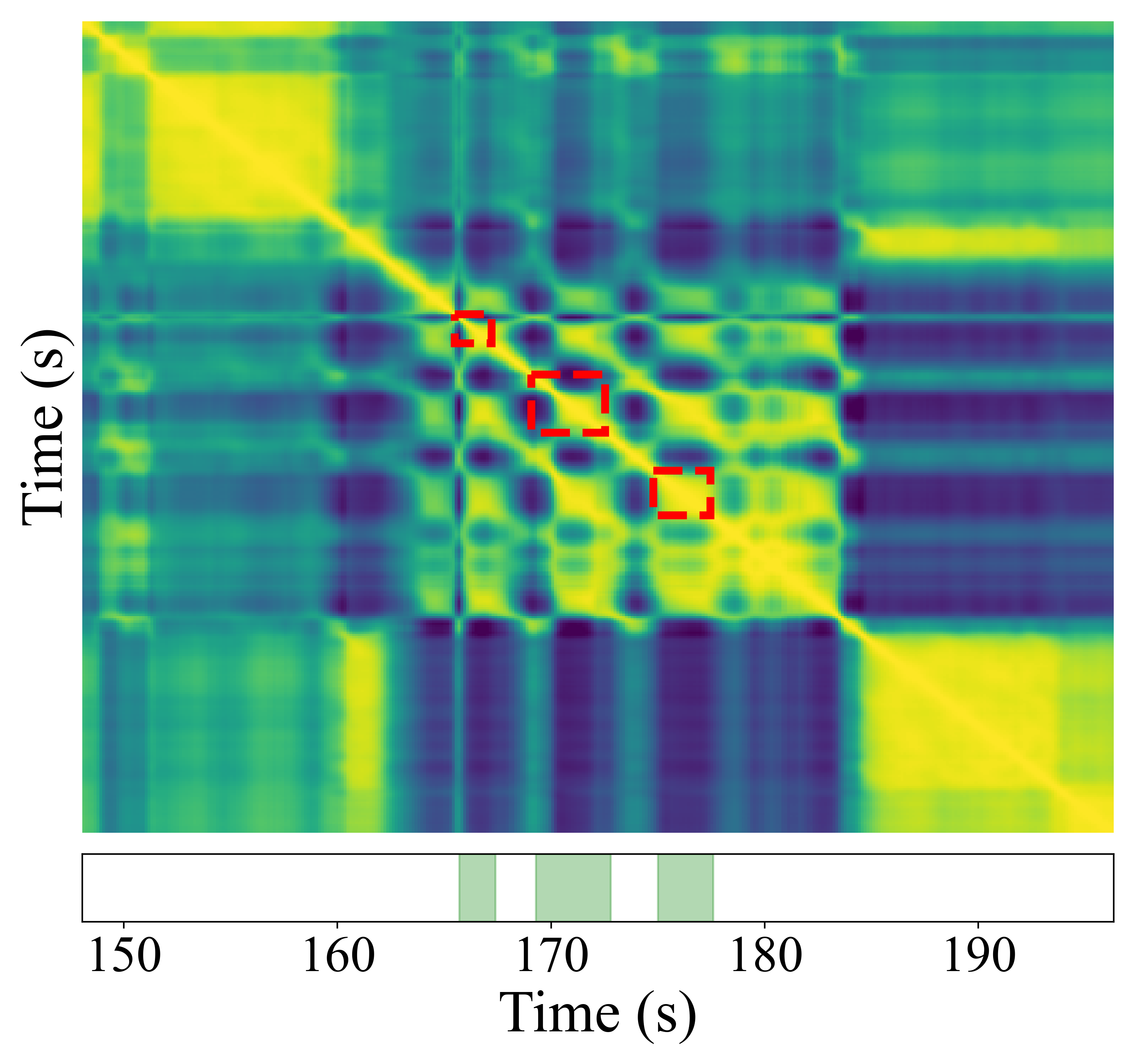}
        \vspace{-15pt}
        \caption{\modelname{}}
        \label{fig:cka_te_tad}
    \end{subfigure}
    \hfill

    \vspace{-9pt}
    \caption{\textbf{Comparison of cosine similarity on encoder output features between TE-TAD and \modelname{}.} Top: represents cosine similarity, with red boxes indicating regions of similarity among features within the ground truth. Bottom: displays the ground truth action timeline for reference. Sample taken from THUMOS14.}

    \vspace{-10pt}
    \label{fig:enc_sim}
\end{figure}
\noindent \textbf{Visualization of Cosine Similarity on Encoder Output Features} 
Fig.~\ref{fig:enc_sim} presents a comparison of cosine similarity matrices for encoder output features between TE-TAD and \modelname{}. Our \modelname{} shows improved feature discriminability, indicating that \encodername{} captures distinct temporal patterns more effectively compared to TE-TAD~\cite{TE-TAD}.

\begin{figure}[!t]
    \centering
    \begin{subfigure}[t]{\linewidth}
        \centering
        \includegraphics[width=0.98\linewidth]{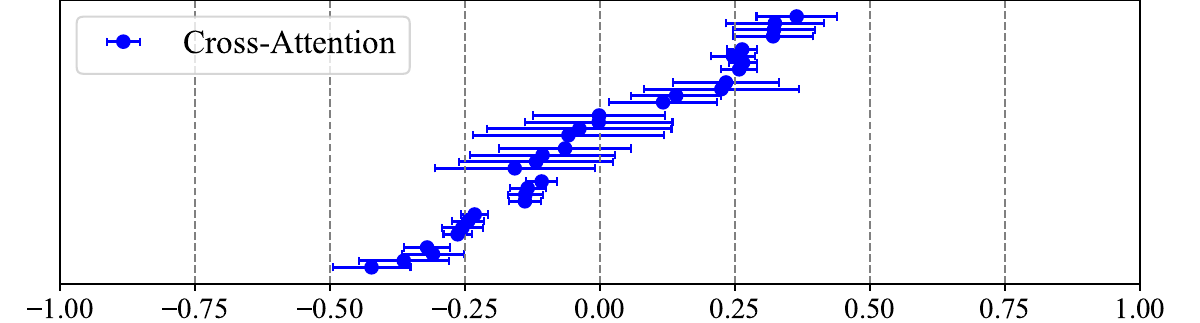}
        \vspace{-1pt}
        \caption{TE-TAD~\cite{TE-TAD}}
    \end{subfigure}%
    \hfill
    \begin{subfigure}[t]{\linewidth}
        \centering
        \includegraphics[width=0.98\linewidth]{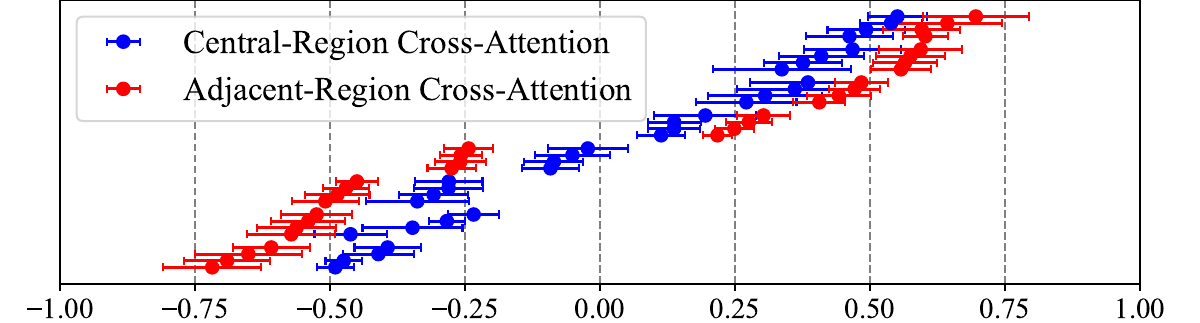}
        \vspace{-1pt}
        \caption{\modelname}
    \end{subfigure}

    \vspace{-8pt}
    \caption{\textbf{Visualization of sampling offsets in cross-attention layers on THUMOS14.} Each point denotes the mean value across the dataset, with error bars indicating the standard deviation.}

    \vspace{-16pt}
    \label{fig:ref}
\end{figure}

\noindent \textbf{Visualization of Sampling Offsets}  
Fig.~\ref{fig:ref} shows the distribution of sampling offsets relative to the center and width of each action query.
The values 0, -1, and 1 correspond to the center $c_q$, the start boundary $c_q - d_q$, and the end boundary $c_q+d_q$, respectively.
This visualization shows that our method gathers information from diverse sampling points, covering both central and adjacent regions.
Furthermore, while deformable attention allows learnable offsets, they remain close to their initial points, indicating the importance of the initial value of $b$ in Eq. \eqref{eq:sampling_offset}.

\section{Conclusion}
\label{sec:conclusion}
In this paper, we propose a multi-dilated gated encoder and central-adjacent region integrated decoder for temporal action detection transformer (\modelname{}).
\encodername{} offers diverse receptive fields while maintaining a single-scale encoding structure by utilizing multi-dilated convolutions.
\decodername{} provides essential information to precisely detect action instances by focusing on both the central- and adjacent- regions of action instances.
Extensive experiments demonstrate that \modelname{} outperforms the previous query-based methods.
Furthermore, our method consistently improves when integrated with the existing query-based detectors.

\newpage
\section*{Acknowledgement}
This work was supported by the Institute of Information \& Communications Technology Planning \& Evaluation (IITP) grant, funded by the Korea government (MSIT) (No. RS-2019-II190079, Artificial Intelligence Graduate School Program (Korea University), No. RS-2024-00457882, AI Research Hub Project, and No. RS-2022-II220984, Development of Artificial Intelligence Technology for Personalized Plug-and-Play Explanation and Verification of Explanation).
{
    \bibliographystyle{ieeenat_fullname}
    \bibliography{main}
}

\end{document}